\documentclass[10pt,twocolumn,letterpaper]{article}

\usepackage{iccv}
\usepackage{times}
\usepackage{epsfig}
\usepackage{graphicx}
\usepackage{amsmath}
\usepackage{amssymb}
\usepackage{booktabs, multirow}
\usepackage[table,xcdraw]{xcolor}
\usepackage[accsupp]{axessibility}
\newcommand{\gr}{\rowcolor[gray]{.95}}
\definecolor{table_bg}{gray}{0.93}  
\graphicspath{{image/}}

\usepackage[breaklinks=true,bookmarks=false]{hyperref}

\iccvfinalcopy 

\def\iccvPaperID{4100} 

\ificcvfinal\fi

\begin{document}

\title{Scene-Aware Feature Matching}
\author{Xiaoyong Lu, Yaping	Yan, Tong Wei, Songlin Du\footnotemark[1]\\
Southeast University, Nanjing, China\\
{\tt\small $\{$luxiaoyong, yan, weit, sdu$\}$@seu.edu.cn}
}

\maketitle
\ificcvfinal\thispagestyle{empty}\fi

\renewcommand{\thefootnote}{\fnsymbol{footnote}}
\footnotetext[1]{Corresponding author.}

\begin{abstract}
Current feature matching methods focus on point-level matching, pursuing better representation learning of individual features, but lacking further understanding of the scene.
This results in significant performance degradation when handling challenging scenes such as scenes with large viewpoint and illumination changes.
To tackle this problem, we propose a novel model named SAM, which applies attentional grouping to guide Scene-Aware feature Matching. 
SAM handles multi-level features, i.e., image tokens and group tokens, with attention layers, and groups the image tokens with the proposed token grouping module.
Our model can be trained by ground-truth matches only and produce reasonable grouping results.
With the sense-aware grouping guidance, SAM is not only more accurate and robust but also more interpretable than conventional feature matching models.
Sufficient experiments on various applications, including homography estimation, pose estimation, and image matching, demonstrate that our model achieves state-of-the-art performance.
\end{abstract}

\section{Introduction}
Feature matching, which refers to finding the correct correspondence between two sets of features, is a fundamental problem for many computer vision tasks, such as object recognition \cite{sift-flow}, structure from motion (SFM) \cite{sfm}, and simultaneous localization and mapping (SLAM) \cite{slam}.
But with illumination changes, viewpoint changes, motion blur and occlusion, it is challenging to find invariance and obtain robust matches from two images.

The classic image matching pipeline generally consists of four parts: (1) feature detection (2) feature description (3) feature matching (4) outlier filtering.
For feature detection, keypoints that have distinguishable features to facilitate matching are detected.
For feature description, descriptors are extracted based on keypoints and their neighborhoods.
The keypoint positions and the corresponding descriptors are employed as features of the image.
Then the feature matching algorithm is applied to find the correct correspondence in two sets of extracted features.
Finally, the outlier filtering algorithm is applied to identify and reject outlier matches based on the obtained matches.

\begin{figure}[t]
	\includegraphics[width=0.99\linewidth]{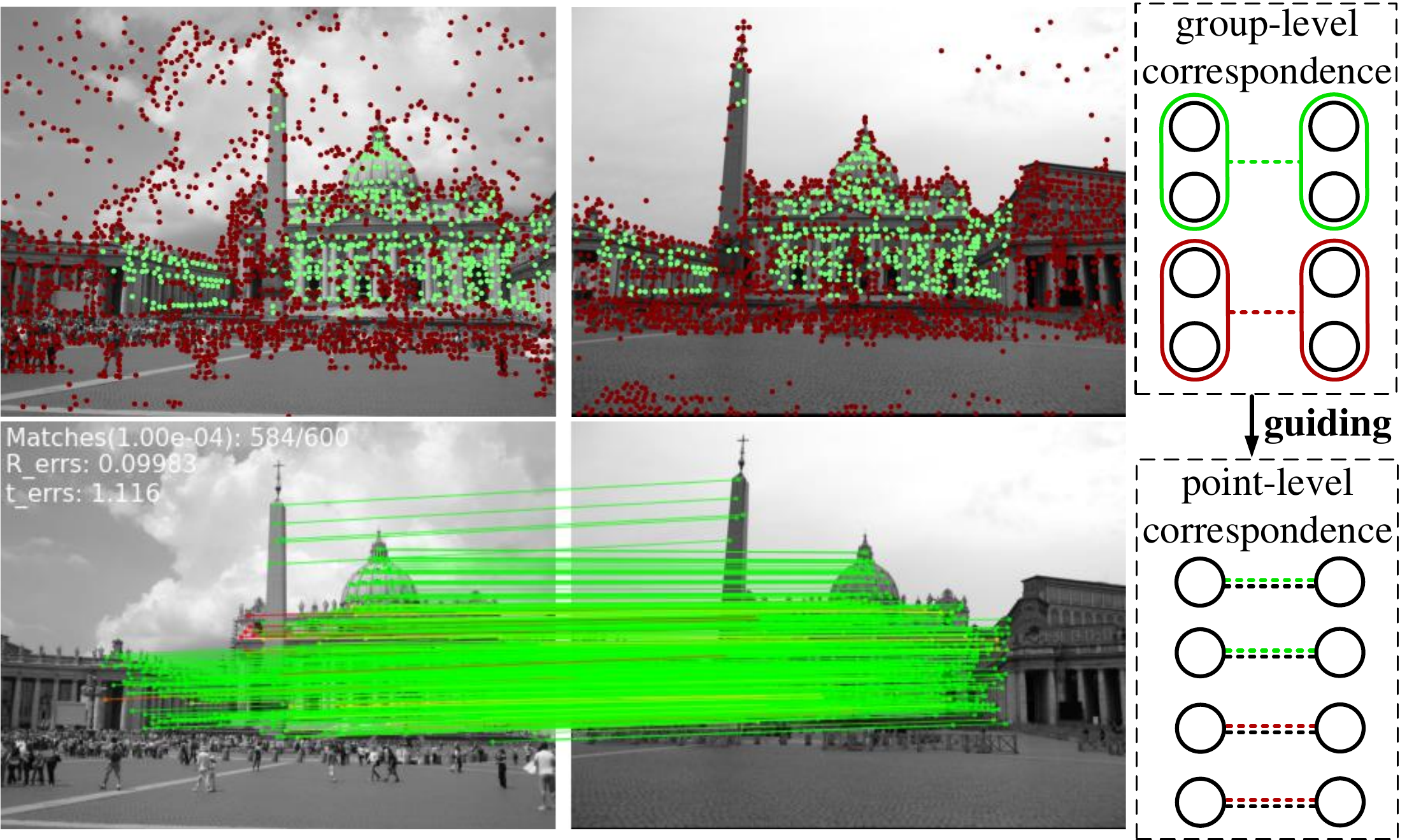}
	\centering
	 \caption{An illustration of the grouping (top) and matching (bottom) result of our proposed method.
	 Points in corresponding regions in the two images are correctly assigned to the same group, 
	 and the grouping information provides scene-aware guidance for point-level feature matching.}
	 \label{figure1}
\end{figure}

The current dominant approaches for image matching are learning-based descriptors with attention-based feature matching models.
Learning-based descriptors extract local features with better representation capabilities by convolutional neural networks (CNN).
Attention-based networks can enhance local features by perceiving global information and modeling contextual relationships between local features.
While the above feature matching pipeline is the dominant method,
the model performance still degrades significantly when dealing with extreme cases, such as scenes with large viewpoint changes and illumination changes.
Because current methods only find correspondences at the low level, i.e., point-level textures, 
and do not incorporate scene-aware information, such as grouping information, semantic information, etc.
Therefore, intra- and inter-image grouping is introduced to SAM to guide point-level attention-based matching.

In this work, we take point-level descriptors as image tokens and additionally introduce the concept of group tokens, which are selected from image tokens by the proposed group token selection module.
Each group token represents a group of features shared by two images, and we intend to assign the corresponding points in both images to the same group, 
while the points that do not correspond to each other are assigned to different groups.
We apply Transformer to model the relationship between image tokens and group tokens for intra- and inter-images, and propose a token grouping module to assign image tokens to different groups based on similarity.
A novel multi-level score strategy is proposed to utilize the scene-aware grouping information to give guidance on point-level features, 
and to obtain reasonable grouping results relying only on ground-truth match supervision.

In summary, the contributions of this paper include:

\begin{itemize}
    \item We propose a novel feature matching model SAM, which allows feature matching to rely on more than point-level textures by introducing group tokens to construct scene-aware features.
    \item The multi-level feature attention encoder and token grouping module are proposed to enable image tokens and group tokens to perceive global context information and assign image tokens to different groups.
    \item We are the first to utilize only ground-truth match supervision to enable the feature matching model to perform the scene-aware grouping and matching through the proposed multi-level score.
    \item SAM achieves state-of-the-art performance in multiple experiments while demonstrating outstanding robustness and interpretability.
\end{itemize}

\section{Related Work}
\textbf{Local Feature Matching.}
For feature detection and description, there are many well-known works on handcrafted methods, such as SIFT \cite{sift}, SURF \cite{surf}, BRIEF \cite{brief} and ORB \cite{orb},
which have good performance and are widely used in 3D computer vision tasks.
With the rise of deep learning, many learning-based detectors have been proposed to further improve the robustness of descriptors under illumination changes and viewpoint changes, 
such as R2D2 \cite{r2d2}, SuperPoint \cite{superpoint}, D2-Net \cite{d2net} and LF-Net \cite{lfnet}.

In addition to detectors, other works have focused on how to get better matches with the obtained local features.
SuperGlue \cite{superglue} is the pioneering work to propose an attention-based feature matching network, 
where the self-attention layer utilizes global contextual information in one image to enhance features 
and the cross-attention layer finds potential matches in two images and performs information interaction on potential matching features.
OETR \cite{oetr} further detects the commonly visible region with an object detection algorithm to limit the attention-based feature matching to the overlap region.
Besides matching the sparse descriptors generated by the detector, LoFTR \cite{loftr} applies self- and cross-attention directly to the feature maps extracted by the convolutional neural network and generates matches in a coarse to fine manner.
MatchFormer \cite{matchformer} further abandons the CNN backbone network and adopts a pure attention architecture, which can perform both feature extraction and feature matching.
There are also methods named cost-volume-based methods \cite{ncnet, dual, 9320451}, which find matches through 4D convolutional neural networks.

The above methods model the relationship between features at the point level and do not utilize higher-level scene information, 
resulting in non-robustness and significant performance degradation when handling large viewpoint changes and illumination changes.
By introducing the concept of group tokens, we group the two image features based on the scene-aware information and
construct multi-level features to make the model more robust when dealing with challenging scenes.

\textbf{Vision Transformer.}
Inspired by the great success of Transformers in the field of Natural Language Processing, researchers have attempted to apply Transformer to Computer Vision.
A. Dosovitskiy et al. \cite{Transformer} first proposed a pure Transformer named ViT directly to sequences of image patches for image classification.
Many variants are subsequently proposed to solve various tasks. 
For feature matching, SuperGlue \cite{superglue} and LoFTR \cite{loftr} are the first sparse and dense matching methods to apply Transformer.
SuperGlue applies classical scaled dot product attention, and LoFTR applies linear attention \cite{linear} to reduce runtime and memory consumption.

Our model is also an application of Transformer for the feature matching task. 
The global perceptual field of the attention mechanism facilitates the local features within and between images to perceive the global contextual information, making the matching more robust.
And attention operations between image tokens and group tokens enable features to learn scene-aware grouping information.

\begin{figure*}[t]
	\includegraphics[width=0.99\linewidth]{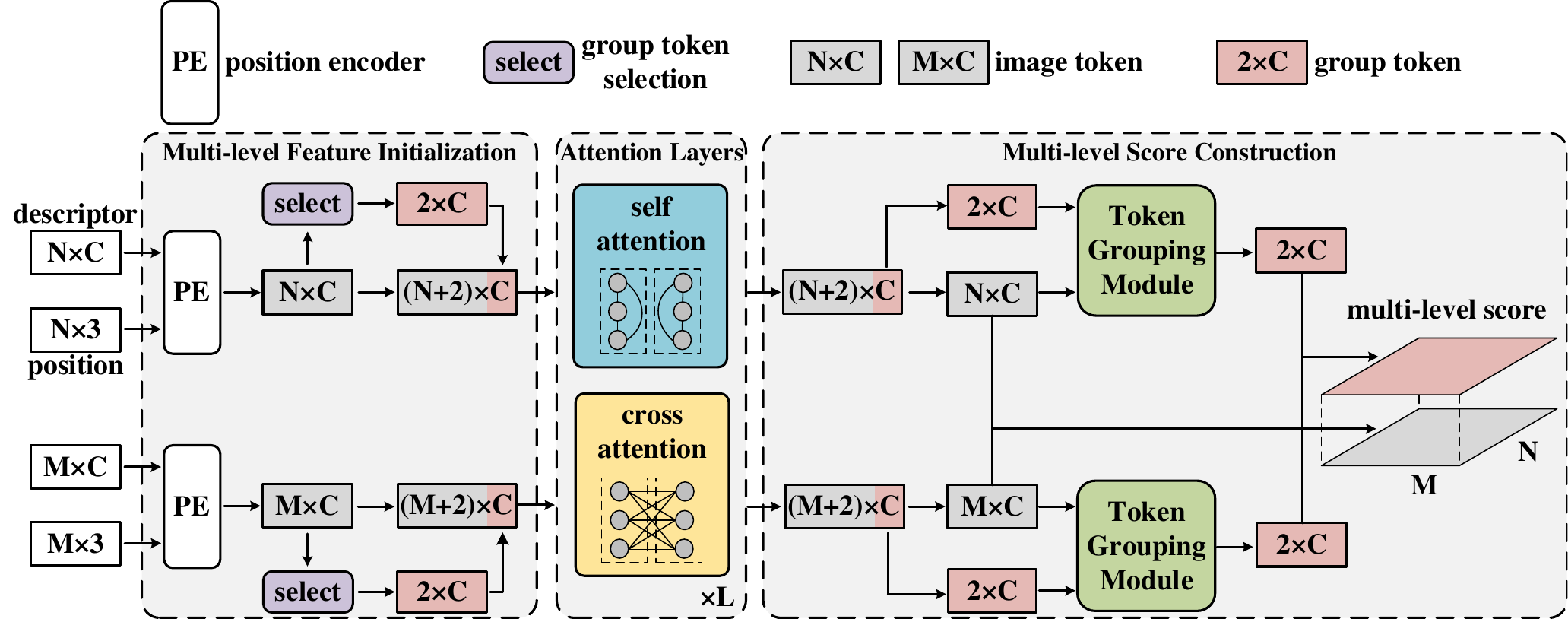}
	\centering
	 \caption{
	 \textbf{Proposed architecture.} SAM consists of three parts, namely multi-level feature initialization, attention layers, and multi-level score construction.
	 SAM first applies the position encoder to obtain position-aware descriptors, i.e. image tokens, and then selects group tokens to form multi-level features with the image tokens.
	 The multi-level features are processed through attention layers, 
	 then the image tokens are assigned to different groups through the token grouping module, 
	 and the grouping information is employed to guide point-level matching by constructing multi-level scores.
	 }
	 \label{figure2}
\end{figure*}

\begin{figure}[t]
	\includegraphics[width=0.85\linewidth]{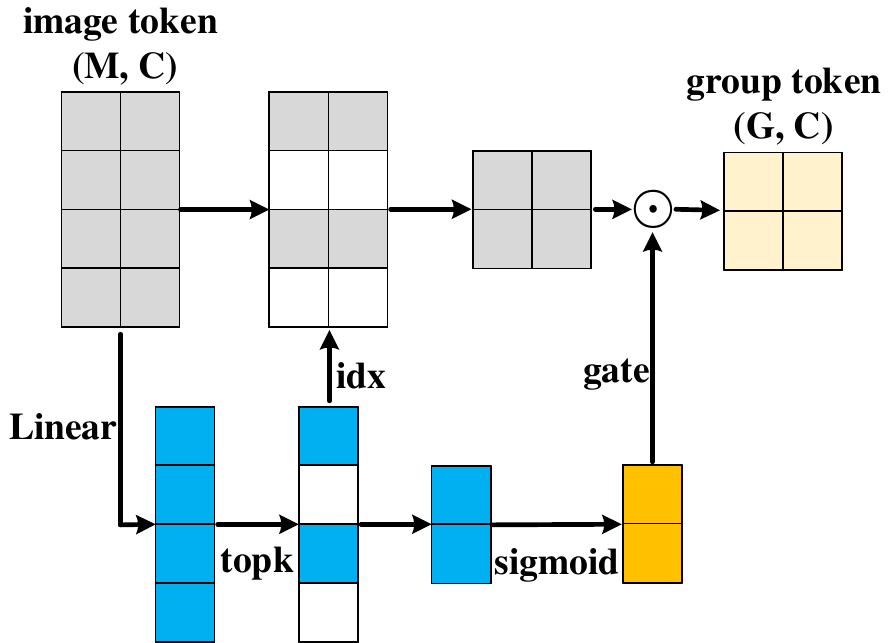}
	\centering
	 \caption{
	\textbf{Group token selection module.} 
	The proper group tokens are chosen from image tokens based on the score projected by image tokens. 
	And the gate signal computed from the score enables the block to be trained end-to-end.
	 }
	 \label{figure3}
\end{figure}

\textbf{Grouping.}
Most learning-based grouping models follow a pipeline of first learning representations with deep neural networks and then applying the grouping algorithms.
For representation learning networks, common types of networks include multi-layer perceptron (MLP) \cite{IMSAT}, CNN \cite{SCCNN} and Variational Autoencoder (VAE) \cite{VaDE}.
Our model, on the other hand, applies Transformer as the representation learning network, globally perceiving image tokens and group tokens to learn deep representations.

For supervision, there are several commonly used grouping losses, 
which are K-means loss \cite{kmeans}, Cluster assignment hardening \cite{hardening}, Cluster classification loss \cite{classification} and Locality-preserving loss \cite{preserving}.
Our grouping algorithm relies only on ground-truth matches to encourage the corresponding points of two images to be assigned to the same group.

\section{Methodology}
Assume that two sets of features $d_s\in\mathbb{R}^{M\times C}$, $d_t\in\mathbb{R}^{N\times C}$ need to be matched which are descriptors extracted from two images.
Subscripts $s$ and $t$ stand for the source image and target image, respectively.
$M$, $N$ are the number of descriptors, and $C$ is the channels of descriptors. 
The keypoint positions are denoted as $p_s\in\mathbb{R}^{M\times 3}$, $p_t\in\mathbb{R}^{N\times 3}$,
which consist of two coordinates and detection confidence.
Our objective is to find the correct correspondence between the two images utilizing descriptors and position information.

An overview of the network architecture is shown in Figure \ref{figure2}.
The position information is first embedded into descriptors by the wave position encoder \cite{paraformer}.
The position-aware descriptors are named image tokens, which are concatenated with the selected group tokens to form multi-level features.
The self-attention layer and cross-attention layer are applied for $L$ times to utilize the global context to enhance multi-level features, 
which are then re-split into image tokens and group tokens. The token grouping module is applied to assign image tokens to different groups based on similarity.
Finally, the multi-level score is constructed from point-level features and group-level features to perform scene-aware matching.

\begin{figure}[t]
	\includegraphics[width=0.95\linewidth]{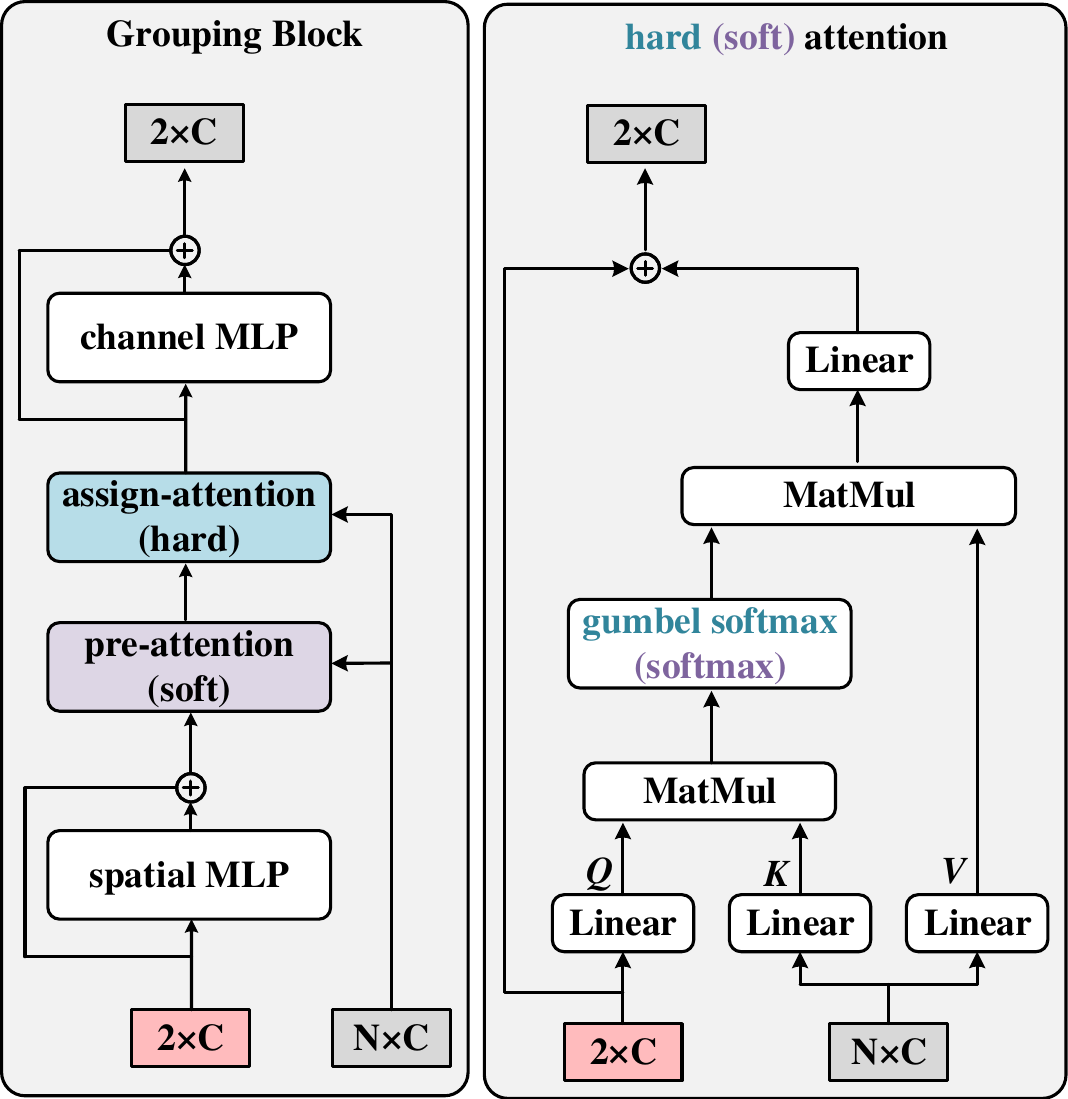}
	\centering
		\caption{\textbf{Token Grouping Module.}
		The pre-attention layer is employed to establish the relationship between image tokens and group tokens before assignment.
		The assign-attention layer is employed to assign image tokens to different groups based on the similarity between image tokens and group tokens.
		The pre-attention and the assign-attention apply $\mathrm{Softmax}$ and $\mathrm{Gumbel}$-$\mathrm{Softmax}$ functions respectively.
		}
		\label{figure4}
\end{figure}

\subsection{Group Token Selection}
The importance of clustering centers has been demonstrated by many clustering algorithms, 
without which the performance of the algorithm degrades or even appears as a trivial solution.
To achieve the best scene-aware grouping effect, the group token selection module is proposed to select the proper group token among the image tokens.

As shown in Figure \ref{figure3}, we first apply a linear layer to compute the image tokens $f$ as group scores $s$, which measure how effective each image token is as a group token.
Then $k$ image tokens with the highest group scores are selected as the group tokens $\tilde{g}$. 
The number of groups $k$ is set to $2$, representing overlapping and non-overlapping regions.
To enable end-to-end training of the group token selection module, we apply the $\mathrm{sigmoid}$ function to calculate the group score $s$ as a gating signal 
and multiply it with the selected image tokens to get the final group tokens $g$.
The group token selection module is denoted as
\begin{equation}
	\begin{split}
		&s  = \mathrm{Linear}(f), \\
		&idx  = \mathrm{rank}(s,k), \\
		&\tilde{g}  = f(idx, :), \\
		&gate  = \mathrm{sigmoid}(s(idx)), \\
		&g  = \tilde{g} \odot gate,
		\label{equ3} 
	\end{split}
\end{equation}

\noindent{where $\odot$ represents the element-wise matrix multiplication.}

\subsection{Multi-level Feature Attention}
To perform information propagation between image tokens and group tokens, we concatenate them,
i.e., $f_s$ and $g_s$, $f_t$ and $g_t$, to form the multi-level features $\hat{f}_s$, $\hat{f}_t$, 
which are processed by the self-attention layer and the cross-attention layer. 
Specifically, the two sets of multi-level features are projected into two sets of $Q, K, V$ by three linear layers, and we compute attention as

\begin{equation}
	\begin{split}
		&\mathcal{SA}_{s,t}  = \mathrm{Softmax}(Q_{s,t}K_{s,t}^{T}/\sqrt{d})V_{s,t}, \\
		&\mathcal{CA}_{s}  = \mathrm{Softmax}(Q_{s}K_{t}^{T}/\sqrt{d})V_{t}, \\
		&\mathcal{CA}_{t}  = \mathrm{Softmax}((Q_{s}K_{t}^{T})^{T}/\sqrt{d})V_{s},
		\label{equ4} 
	\end{split}
\end{equation}

\noindent{where $\mathcal{SA}$ and $\mathcal{CA}$ denote self-attention and cross-attention output, and $d$ is the feature dimension.} The inputs of self-attention come from the feature of one image, 
such as $(Q_{s}, K_{s}, V_{s})$ or $(Q_{t}, K_{t}, V_{t})$, 
while the inputs of the cross-attention come from features of different images. 
To save computational costs and memory consumption, $Q_{t}K_{s}^{T}$ is replaced by $(Q_{s}K_{t}^{T})^{T}$, since $Q_{t}K_{s}^{T}$ and $Q_{s}K_{t}^{T}$ are highly correlated.

Two MLPs are applied to adaptively fuse the self- and cross-attention outputs, 
and the fusion outputs are used to update the features as the input of the next attention layer.

\begin{equation}
	\begin{split}
		&\hat{f}_{s}^{l+1}  = \hat{f}_{s}^{l}+\mathrm{MLP}_{s}([\hat{f}_{s}^{l}|\mathcal{SA}_{s}^{l}|\mathcal{CA}_{s}^{l}]), \\
		&\hat{f}_{t}^{l+1}  = \hat{f}_{t}^{l}+\mathrm{MLP}_{t}([\hat{f}_{t}^{l}|\mathcal{SA}_{t}^{l}|\mathcal{CA}_{t}^{l}]).
		\label{equ5} 
	\end{split}
\end{equation}

We stack $L=9$ multi-level feature attention layers, which model the relationship between image tokens, 
and between image tokens and group tokens of two sets of features.
Attention between image tokens finds potential matches at the point level between features, 
while attention between group tokens and image tokens allows group tokens to perceive global context and facilitates subsequent token grouping module.

\subsection{Token Grouping Module}
As shown in Figure \ref{figure4}, to assign image tokens to different groups based on similarity with group tokens in the embedding space,
four parts are employed to form the token grouping module, namely spatial MLP, pre-attention layer, assign-attention layer and channel MLP.

Spatial MLP and channel MLP are introduced to the token grouping module to enhance the module capacity,
which contain two fully-connected layers and an element-wise nonlinearity.
Specifically, spatial MLP is applied to the transposed input $g^{T}\in\mathbb{R}^{C\times 2}$, mapping $\mathbb{R}_{2} \mapsto \mathbb{R}_{2}$ to interact between group tokens. 
And channel MLP is applied to $g\in\mathbb{R}^{2\times C}$, mapping $\mathbb{R}_{C} \mapsto \mathbb{R}_{C}$ to interact between channels.

\begin{equation}
	\begin{split}
		&\mathrm{spatial\;MLP}: O_{*,i}=I_{*,i}+W^{s}_{2}\sigma(W^{s}_{1}I_{*,i}), \\
		&\mathrm{channel\;MLP}: O_{j,*}=I_{j,*}+W^{c}_{2}\sigma(W^{c}_{1}I_{j,*}),
		\label{equ6} 
	\end{split}
\end{equation}

where $W^{s}_{1}$, $W^{s}_{2}, W^{c}_{1}$, $W^{c}_{2}$ are learnable weight matrices, and $\sigma$ is the $\mathrm{GELU}$ function.

As core parts of the token grouping module, 
the pre-attention layer and assign-attention layer perform the information propagation between image tokens and group tokens,
and assign image tokens based on similarity, respectively.
Both pre-attention layer and assign-attention layer 
apply three linear layers projecting group tokens as $Q$ and image tokens as $K, V$.
The difference between the two attention is that $\mathrm{Softmax}$ is applied for pre-attention layer, 
and $\mathrm{Gumbel}$-$\mathrm{Softmax}$ is applied for assign-attention layer. 

Soft attention weight $A$ for pre-attention layer and assign attention weight $\tilde{A}$ for assign-attention layer are denoted as

\begin{equation}
	\begin{split}
		&A = \mathrm{Softmax}(QK^{T}) , \\
		&\tilde{A}_{i,j}=\frac{\mathrm{exp}(Q_{i}K_{j}^{T}+G_{i})}{\sum_{k=1}^{2}\mathrm{exp}(Q_{k}K^{T}_{j}+G_{k})}.
		\label{equ7} 
	\end{split}
\end{equation}

\noindent{$G$ are i.i.d. samples drawn from $\mathrm{Gumbel}(0,1)$ distribution.}

After getting the assign attention weight, we decide the group to which the image tokens are assigned by the $\mathrm{argmax}$ over all group tokens.
Since the $\mathrm{argmax}$ operation is not differentiable, the straight-through trick in \cite{GroupViT} is applied to compute the assignment matrix as

\begin{equation}
	\begin{split}
		&\hat{A}=\tilde{A}_{argmax}+\tilde{A}-\mathrm{sg}(\tilde{A}),
		\label{equ8} 
	\end{split}
\end{equation}

\noindent{where $\mathrm{sg}(\cdot)$ is the stop gradient operation. By the straight-through trick, assignment matrix $\hat{A}$ is numerically equal to the one-hot matrix $\tilde{A}_{argmax}$,
and the gradient of $\hat{A}$ is equal to the gradient of assign attention weight $\tilde{A}$. 
Based on the assignment matrix $\hat{A}$, all the image tokens of the same group are weighted summed to update the group tokens.}

\subsection{Multi-level Score}
Conventional feature matching methods compute the dot product of two sets of features as the point-level score matrix, and then select matches based on it.
We compute both point-level score matrix $S^{f}\in\mathbb{R}^{M\times N}$ and group-level score matrix $S^{g}\in\mathbb{R}^{2\times 2}$ for the two sets of features based on image tokens $f$ and group tokens $g$.

\begin{equation}
	\begin{split}
		&S^{f}_{i,j}=<f^{s}_{i},f^{t}_{j}>, \\
		&S^{g}_{i,j}=<g^{s}_{i},g^{t}_{j}>.
		\label{equ9} 
	\end{split}
\end{equation}

To utilize the group information to guide the point-level matching, 
the group-level score matrix is expanded to point-level $\hat{S}^{g}\in\mathbb{R}^{M\times N}$ with the soft attention weights $A_{s}\in\mathbb{R}^{M\times 2}, A_{t}\in\mathbb{R}^{N\times 2}$ of the two sets of features.
The point-level score and the group-level score are weighted summed to obtain the multi-level score matrix $S$,

\begin{equation}
	\begin{split}
		&\hat{S}^{g}=A_{s}S^{g}A_{t}^{T}, \\
		&S=\alpha S^{f}+\beta\hat{S}^{g},
		\label{equ10} 
	\end{split}
\end{equation}

\noindent{where $\alpha$ and $\beta$ are learnable parameters.}
The multi-level score matrix $S$ is taken as the cost matrix of the optimal transport problem. 
The Sinkhorn algorithm \cite{Sinkhorn} is applied to iteratively obtain the optimal partial assignment matrix $P$. 
Based on $P$, matches with $P_{ij}$ less than the matching threshold $\theta$ are filtered first, then the mutual nearest neighbor criterion is employed to select the final matches $M$.

For supervision, ground-truth matches \emph{$M_{gt}$} and non-matching point sets $I, J$ are computed from homography or camera pose and depth map.
The first loss is the negative log-likelihood loss $Loss_{m}$ over the optimal partial assignment matrix $P$,
which is denoted as

\begin{equation}
	\begin{split}
		Loss_{m} = &- \frac{1}{|M_{gt}|}\sum_{(i,j)\in M_{gt}} \log P_{i,j} \\
		&-\frac{1}{|I|}\sum_{i\in I} \log P_{i,M+1} - \frac{1}{|J|}\sum_{j\in J} \log P_{N+1,j}.
		\label{equ11} 
	\end{split}
\end{equation}
$P_{*, M+1}$ and $P_{N+1, *}$ are learnable dustbins to which non-matching points are assigned.
$Loss_{m}$ provides supervision for matching and implicit supervision for grouping,
as it encourages the assignment of corresponding points in two images to the same group,
and the non-corresponding points to a different group.

To further enhance the ability to group precisely, we apply explicit supervision to assign attention weight $\tilde{A}$ with ground-truth point sets $M_{gt}$, $I$ and $J$.
\begin{equation}
	\begin{split}
		Loss_{g} = &- \frac{1}{|M_{gt}|}\sum_{(i,j)\in M_{gt}} (\log \tilde{A}^{s}_{i,0} + \log \tilde{A}^{t}_{j,0}) \\
		&-\frac{1}{|I|}\sum_{i\in I} \log \tilde{A}^{s}_{i,1} - \frac{1}{|J|}\sum_{j\in J} \log \tilde{A}^{t}_{j,1}.
		\label{equ12} 
	\end{split}
\end{equation}

\section{Experiments}
\subsection{Implementation Details}
SAM is trained on the Oxford100k dataset \cite{100k} for homography estimation experiments
and on the MegaDepth dataset \cite{MegaDepth} for pose estimation and image matching experiments.
Our PyTorch implementation of SAM involves $L=9$ attention layers, and all intermediate features have the same dimension $C=256$.
The matching threshold $\theta$ is set to 0.2.
For the homography estimation experiment, we employ the AdamW \cite{adam} optimizer for 10 epochs using the cosine decay learning rate scheduler and 1 epoch of linear warm-up.
A batch size of 8 and an initial learning rate of 0.0001 are used.
For the outdoor pose estimation experiment, we use the same AdamW optimizer for 30 epochs using the same learning rate scheduler and linear warm-up.
A batch size of 2 and an initial learning rate of 0.0001 are used.
All experiments are conducted on a single NVIDIA GeForce RTX 2060 SUPER GPU, 16GB RAM and Intel Core i7-10700K CPU.

\subsection{Homography Estimation}
\noindent\textbf{Dataset.}
For the homography estimation, SAM is trained on the Oxford100k dataset \cite{100k}
and evaluated on the $\mathcal{R}$1M dataset \cite{r1m}.
To perform self-supervised training, 
we randomly sample ground-truth homography by limited parameters to generate image pairs.
We resize images to 640$\times$480 pixels and detect 512 keypoints in the image.
When the detected keypoints are not enough, random keypoints are added for efficient batching.

\noindent\textbf{Baselines.}
We employ the Nearest Neighbor (NN) matcher, NN matcher with outlier filtering methods \cite{pointcn,oanet}, and attention-based matcher SuperGlue \cite{superglue} as baseline matchers.
All matchers in Table \ref{figure1} apply SuperPoint \cite{superpoint} as input descriptors for a fair comparison.
The results of SuperGlue are from our implementation.

\noindent\textbf{Metrics.}
The ground-truth matches are computed from the generated homography and the keypoint coordinates of the two images.
A match is considered correct if the reprojection error is less than $3$ pixels. 
We evaluate the precision and recall based on the ground-truth matches and compute the F1-score.
We calculate reprojection error with the estimated homography and report the area under the cumulative error curve (AUC) up to 10 pixels.

\noindent\textbf{Results.} 
As shown in Table \ref{table1}, SAM achieved the best performance in the homography estimation experiment. 
Benefiting from the powerful modeling capability of Transformer, the attention-based matcher is significantly ahead of other methods.
Compared to the state-of-the-art outlier filtering method OANet, SAM achieves a $+21\%$ lead on F1-score.
Both as attention-based methods, SAM is ahead of SuperGlue in both precision and recall because grouping information is introduced in addition to point-level matching, 
eliminating unreasonable matches and strengthening matches in the same group based on the grouping information.
It ends up with a $+1.92\%$ advantage over SuperGlue on the F1-score. Qualitative results of matching and grouping are shown in Figure \ref{figure6}.

\begin{table}[t]
	\centering
	\resizebox{.9\columnwidth}{!}{
	\begin{tabular}{@{}lcccc@{}}
	\toprule
		Matcher                                       & AUC                                          & Precision                                       & Recall                                   & F1-score           \\ \midrule
		NN                                            & 39.47                                         & 21.7                                            & 65.4                                     & 32.59                     \\
		NN + mutual                                   & 42.45                                         & 43.8                                            & 56.5                                     & 49.35    \\
		NN + PointCN                                  & 43.02                                         & 76.2                                            & 64.2                                     & 69.69    \\
		NN + OANet                                    & 44.55                                         & 82.8                                            & 64.7                                     & 72.64    \\
		SuperGlue                                     & 51.94        								  & 86.2    									    & 98.0 								       & 91.72     \\
		\gr
		\textbf{SAM}         						& \textbf{53.80}								& \textbf{89.54}		  						  & \textbf{98.13}     		 				  & \textbf{93.64} \\ \bottomrule  
	\end{tabular}
	}
	\caption{\textbf{Homography estimation on $\mathcal{R}$1M dataset.} AUC @10 pixels is reported. The best method is highlighted in \textbf{bold}.}
	\label{table1}
\end{table}

\subsection{Outdoor Pose Estimation}
\noindent\textbf{Dataset.}
For the outdoor pose estimation experiment, the model is trained on the MegaDepth dataset \cite{MegaDepth} and evaluated on the YFCC100M dataset \cite{2016YFCC100M}.
For training, 200 pairs of images in each scene are randomly sampled for each epoch. For evaluation,
the YFCC100M image pairs and ground truth poses provided by SuperGlue are used. 
For training on the MegaDepth dataset, we resize the images to 960$\times$720 pixels and detect 1024 keypoints.

\noindent\textbf{Baselines.}
The baseline method contains NN matchers with outlier filtering method \cite{oanet} and attention-based matcher SuperGlue \cite{superglue}.
All matchers in Table \ref{figure2} apply SuperPoint \cite{superpoint} as input descriptors for a fair comparison.
The results of SuperGlue are from our implementation.

\noindent\textbf{Metrics.}
The AUC of pose errors at the thresholds ($5^\circ$, $10^\circ$, $20^\circ$) are reported.
Both approximate AUC \cite{oanet} and exact AUC \cite{superglue} are evaluated for a fair comparison.

\noindent\textbf{Results.} 
As shown in Table \ref{table2}, for the outdoor pose estimation experiments, SAM achieves the best performance at all thresholds in both approximate AUC and exact AUC, demonstrating the robustness of our models.
Compared to the attention-based matcher SuperGlue, which only considers point-level matching, our model can bring $(+3.00\%, +3.67\%, +3.12\%)$ improvement on exact AUC
and $(+4.98\%, +4.12\%, +3.45\%)$ improvement on approximate AUC at three thresholds of $(5^{\circ}, 10^{\circ}, 20^{\circ})$, respectively.
In outdoor scenes where large viewpoint changes and occlusions often occur, 
SAM provides scene understanding information for feature matching by utilizing grouping information to block out irrelevant interference.

\begin{figure*}[t]
	\includegraphics[width=0.99\linewidth]{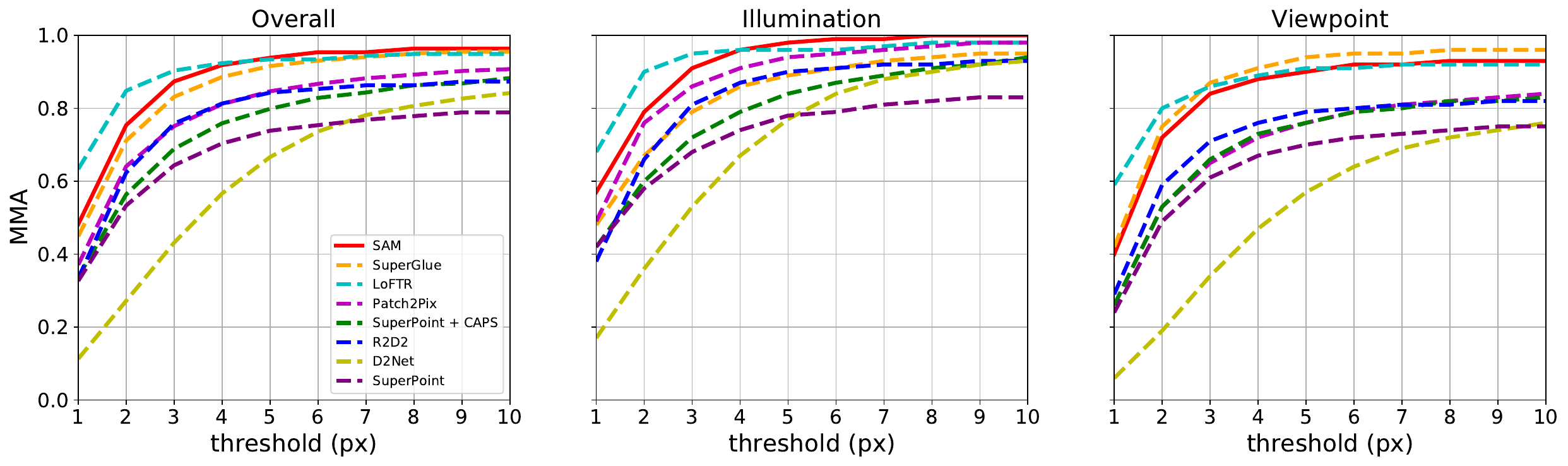}
	\centering
	 \caption{\textbf{Image Matching on HPatches dataset.} MMA curves are plotted by changing the reprojection error threshold. 
	 Our model achieves the best overall performance at reprojection error thresholds of 5 and above.}
	 \label{figure5}
\end{figure*}

\begin{table}[t]
	\centering
	\resizebox{.99\columnwidth}{!}{
	\begin{tabular}{@{}lcccccc@{}}
	\toprule
	\multirow{2.5}{*}{Matcher} & \multicolumn{3}{c}{Exact AUC} & \multicolumn{3}{c}{Approx. AUC} \\ \cmidrule(l){2-7} 
							 & @5$^{\circ}$       & @10$^{\circ}$      & @20$^{\circ}$     & @5$^{\circ}$        & @10$^{\circ}$      & @20$^{\circ}$       \\ \midrule
	NN + mutual              & 16.94    & 30.39    & 45.72   & 35.00     & 43.12    & 54.25    \\
	NN + OANet               & 26.82    & 45.04    & 62.17   & 50.94     & 61.41    & 71.77    \\
	SuperGlue                & 28.45    & 48.6     & 67.19   & 55.67     & 66.83    & 74.58    \\
	\rowcolor{table_bg}
	\textbf{SAM}            & \textbf{31.45}    & \textbf{52.27}    & \textbf{70.31}   & \textbf{60.65}     & \textbf{70.95}    & \textbf{78.03}    \\ \bottomrule
	\end{tabular}%
	}
	\caption{\textbf{Pose estimation on YFCC100M dataset.} Our model lead other methods at all thresholds.}
	\label{table2}
\end{table}

\subsection{Image Matching}
\noindent\textbf{Dataset.}
For the image matching experiment, the same model in the outdoor experiment is used.
We follow the evaluation protocol as in D2-Net \cite{d2net} and evaluate models on 108 HPatches \cite{hpatches} sequences, 
which contain 52 sequences with illumination changes and 56 sequences with viewpoint changes.

\noindent\textbf{Baselines.}
Our model is compared with learning-based descriptors SuperPoint \cite{superpoint}, D2Net \cite{d2net}, R2D2 \cite{r2d2} 
and advanced matchers SuperGlue \cite{superglue}, LoFTR \cite{loftr}, Patch2Pix \cite{patch2pix}, and CAPS \cite{caps}.

\noindent\textbf{Metrics.}
We compute the reprojection error from the ground truth homography and vary the matching threshold to plot the mean matching accuracy (MMA) curve, 
which is the average percentage of correct matches for each image.

\noindent\textbf{Results.} 
As shown in Figure \ref{figure5}, SAM achieves the best overall performance at reprojection error thresholds of 5 and above, 
demonstrating the robustness of our model in response to illumination changes and viewpoint changes.
Experiments find that detector-based matching methods such as SuperGlue work well in viewpoint change scenes, 
while detector-free matching methods such as LoFTR work well in illumination change scenes.
SAM performs well as a detector-based method in illumination change scenes, achieving an MMA of $100\%$ at reprojection error thresholds of 8 and above, 
ahead of the detector-free matchers LoFTR and Patch2Pix and substantially ahead of the detector-based matcher SuperGlue.
For viewpoint change scenes, our model is ahead of LoFTR at error thresholds of 6 and above.

\begin{table}[t]
	\centering
	\resizebox{0.85\linewidth}{!}{%
	\begin{tabular}{@{}lccc@{}}
	\toprule
	Methods  & Precision & Recall & F1-score \\ \midrule
	(i)  random      & 80.88          & 92.48      & 86.29          \\
	(ii) learnable parameters       & 81.22     & 95.10  & 87.61    \\
	\rowcolor{table_bg}
	(iii) \textbf{selection}       & \textbf{81.74}     & \textbf{96.36}  & \textbf{88.45}     \\ \bottomrule
	\end{tabular}
	}
	\caption{\textbf{Ablation study on group token.}}
	\label{table3}
\end{table}

\begin{table}[t]
	\centering
	\resizebox{0.85\linewidth}{!}{%
	\begin{tabular}{@{}lccc@{}}
	\toprule
	Methods  & Precision & Recall & F1-score \\ \midrule
	(i) w/o pre-attention      & 80.85     & 95.84  & 87.71    \\
	(ii)  w/o spatial MLP      & 81.35     & 96.16  & 88.14    \\
	(iii) w soft attention    & 81.60     & 96.15  & 88.28     \\
	(iv) w/o channel MLP      & 81.69     & 96.05  & 88.29    \\
	\rowcolor{table_bg}
	(v) \textbf{full}        & \textbf{81.74}     & \textbf{96.36}  & \textbf{88.45}     \\ \bottomrule
	\end{tabular}
	}
	\caption{\textbf{Ablation study on token grouping module.}}
	\label{table4}
\end{table}

\subsection{Ablation Study}

Comprehensive ablation experiments are conducted on the Oxford100k dataset to prove the validity of our designs.

\noindent\textbf{Group Token Selection.} 
As shown in Table \ref{table3}, we conduct ablation experiments on the methods of generating group tokens, containing random selection, learnable parameters as group tokens, and group token selection module.
Firstly, as methods that can be learned end-to-end, learnable parameters group tokens and selection module outperform unlearnable random selection method, 
demonstrating the importance of proper group tokens.
The token selection module performs better than the learnable parameter tokens because the token selection module can be adaptively selected based on the input image token 
whereas the learnable parameter tokens are static for all the images, 
so learnable parameter tokens have weaker generalizability when dealing with diverse scenes such as outdoor, indoor scenes.

\begin{table}[t]
	\centering
	\resizebox{0.85\linewidth}{!}{%
	\begin{tabular}{@{}lccc@{}}
	\toprule
	Methods  & Precision & Recall & F1-score \\ \midrule
	(i)  w/o group-level score      & 79.95          & 93.79      & 86.31         \\
	\rowcolor{table_bg}
	(ii) \textbf{w group-level score}       & \textbf{81.74}     & \textbf{96.36}  & \textbf{88.45}     \\ \bottomrule
	\end{tabular}
	}
	\caption{\textbf{Ablation study on multi-level score.}}
	\label{table5}
\end{table}

\begin{table}[t]
	\centering
	\resizebox{0.99\linewidth}{!}{%
	\begin{tabular}{@{}lcc@{}}
	\toprule
	Methods  & \#Params (M) & Runtime (ms)  \\ \midrule
	SuperGlue $@2048$ keypoints    & 12.0     & 104.37               \\
    LoFTR $@800 \times 800$ resolution        & 11.6     & 373.50      \\
	ASpanFormer $@800 \times 800$ resolution  & 15.8     & 500.28      \\
	\rowcolor{table_bg}
	SAM $@2048$ keypoints    & 14.8     & 111.42      \\ \bottomrule
	\end{tabular}
	}
	\caption{\textbf{Efficiency analysis.}}
	\label{table6}
\end{table}

\begin{figure*}[t]
	\includegraphics[width=0.99\linewidth]{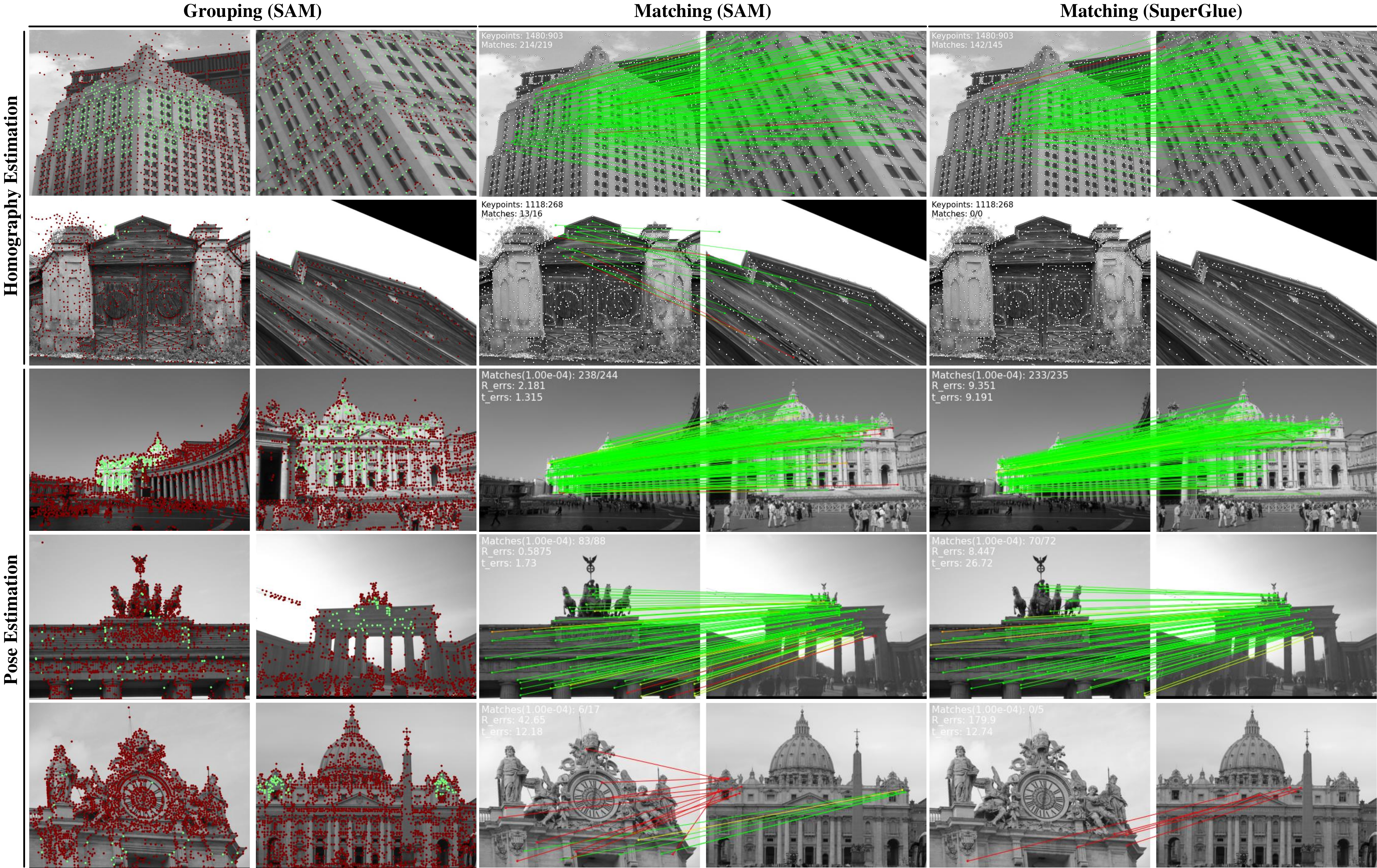}
	\centering
	 \caption{\textbf{Qualitative results for grouping and matching.} SAM can effectively assign the corresponding points of two images into the same group when dealing with 
	 indoor and outdoor scenes, thus guiding a more robust and accurate point-level matching.}
	 \label{figure6}
\end{figure*}

\noindent\textbf{Token Grouping Module.} 
As shown in Table \ref{table4}, without the spatial MLP or channel MLP, the model performance degrades because MLP provides non-linearity and higher dimensional mapping enabling larger model capacity.
And the performance also degrades without the pre-attention layer because the image tokens and the group tokens lose information propagation before assignment.
When replacing hard attention with soft attention in the assign-attention, that is, replacing the one-hot assignment matrix with $A$ obtained by $\mathrm{Softmax}$,
the performance of the token grouping module degrades because soft attention introduces more ambiguity to the group token while hard assignment makes the group tokens more separated.
Stronger explicitness and weaker ambiguity in grouping make each group contain less information about weakly correlated image tokens.

\noindent\textbf{Multi-level Score.} 
Experiments on the multi-level score are conducted to demonstrate the guidance of group-level score to point-level matching.
When the group-level term in the multi-level score is removed, our matcher degenerates to a conventional point-level matcher,
and the performance degrades because the model loses the scene-aware information brought by grouping, demonstrating the effectiveness of our multi-level score design.
Without the group-level term in score, the token grouping module also loses the supervision brought by ground-truth matches, 
so the token grouping module cannot be trained to produce reasonable grouping results. 
\section{Limitation and Discussion}
Firstly, due to the additional group tokens and token grouping module, 
the computational complexity of SAM increases compared to point-level matchers, 
but only two groups do not harm the real-time capability of our model.

Since the token grouping module only utilizes ground-truth matching supervision, the trained model explicitly assigns overlapping regions in two images to one group.
The model is unable to recognize the semantics of buildings or objects in the two scenes to guide matching. 
We believe that the model is able to group more complex semantic information if appropriate supervision is provided.
\section{Conclusion}
In this paper, we present a novel attention-based matcher SAM, which incorporates scene-aware group guidance.
Compared to matching features at the point level, we introduce group tokens on the basis of the image tokens.
Group tokens and image tokens are concatenated to model the global relationship through attention layers, 
and the token grouping module assigns image tokens to scene-aware groups.
We build multi-level score which utilizes point-level and group-level information to generate matching.
Benefiting from the scene-aware information, our model achieves state-of-the-art performance.
SAM is also more interpretable than current feature matching models since grouping information can be visualized.
\section{Acknowledgments}
The authors would like to thank Prof. Min-Ling Zhang for insightful suggestions and fruitful discussions. 
This work was jointly supported by the National Natural Science Foundation of China under grants 62001110 and 62201142, 
the Natural Science Foundation of Jiangsu Province under grant BK20200353, 
and the Shenzhen Science and Technology Program under grant JCYJ20220530160415035.

{\small
\bibliographystyle{ieee_fullname}
\bibliography{egbib}
}

\end{document}